\documentclass{article}
\usepackage{cite}
\usepackage{amsmath,amssymb,amsfonts}
\usepackage{algorithmic}
\usepackage{graphicx}
\usepackage{textcomp}
\usepackage{xcolor}
\usepackage{authblk}
\usepackage[margin=3cm]{geometry}

\usepackage{booktabs}          
\usepackage{multirow}          
\usepackage{amsfonts}         
\usepackage{graphicx} 
\usepackage{duckuments} 
\usepackage[T1]{fontenc}
\usepackage{url}
\usepackage{amsmath}
\usepackage{amssymb}
\usepackage{stfloats}
\usepackage[ruled,linesnumbered]{algorithm2e}
\usepackage[mathscr]{euscript}
\usepackage{comment}
\usepackage{placeins}
\usepackage{relsize}
\usepackage{xspace}
\usepackage{xcolor}
\usepackage[framemethod=TikZ]{mdframed}
\newcommand{\spara}[1]{\smallskip\noindent{\bf #1}}
\newcommand{\method}{\textsf{G{\smaller RAPH}S{\smaller HAP}}\xspace}
\newcommand{\gsfunc}{\xi}

\usepackage{subcaption}
\usepackage{hyperref}
    
\begin{document}

\title{Explaining Identity-aware Graph Classifiers\\ through the Language of Motifs\\
}

\author[1]{Alan Perotti}
\author[1]{Paolo Bajardi}
\author[1,2]{Francesco Bonchi}
\author[1]{André Panisson}
\affil[1]{CENTAI Institute, Turin, Italy}
\affil[2]{Eurecat, Barcelona, Spain}

\date{}

\maketitle 

\begin{abstract}
Most methods for explaining black-box classifiers  (e.g., on tabular data, images, or time series) rely on measuring the impact that removing/perturbing features has on the model output. This forces the explanation language to match the classifier's feature space. However, when dealing with graph data, in which the basic features correspond to the edges describing the graph structure, this matching between features space and explanation language might not be appropriate. 
%
Decoupling the feature space (edges) from a desired high-level explanation language (such as motifs) is thus a major challenge towards developing actionable explanations for graph classification tasks.

In this paper we introduce \method, a Shapley-based approach able to provide motif-based explanations for  identity-aware graph classifiers, assuming no knowledge whatsoever about the model or its training data: the only requirement is that the classifier can be queried as a black-box at will.
For the sake of computational efficiency we explore a progressive approximation strategy and show how a simple kernel can efficiently approximate explanation scores, thus allowing \method to scale on scenarios with a large explanation space (i.e., large number of motifs).
We showcase \method on a real-world brain-network dataset consisting of patients affected by Autism Spectrum Disorder and a control group. Our experiments highlight how the classification provided by a black-box model can be effectively explained by few connectomics patterns.\\


\end{abstract}


\section{Introduction}
\label{sec:intro}
Model interpretability of prediction tasks is increasingly needed for accountability, trust, fairness and debugging~\cite{shneiderman2020human,bhatt2020explainable,panigutti2020fairlens}, especially when machine learning algorithms are incorporated into decision-making processes that may directly affect people’s lives~\cite{tan2018distill}.
Therefore, researchers 
devised techniques to enrich model predictions with explanations~\cite{guidotti2018survey,gilpin2018explaining}. While such methods mainly focus on tabular data~\cite{ribeiro2016should,lundberg2017unified}, and have been further extended to images~\cite{buhrmester2021analysis} and time series~\cite{assaf2019explainable}, much less attention have been devoted to provide actionable explanations for graph classification tasks.
Graphs are a powerful mathematical representation of several real-world systems and, as learning from networked-structures requires ad-hoc approaches~\cite{zhou2020graph}, so do explanation techniques aimed at interpreting algorithmic decisions on graphs.
Most model-agnostic explainability methods 
rely on some form of vectorisation of the original input space: explanations are then provided by ranking the input features according to some score~\cite{ribeiro2016should,lundberg2017unified,chen2018learning}, thus \emph{constraining the explanations to be expressed in the very same language of the training data of the black-box}.
However, when dealing with graph data, in which the basic features correspond essentially to the adjacency information describing the graph structure (i.e., the edges), this matching between features space and explanation language might not be appropriate. In this regard, we argue that $(i)$ a good explanation method for graph classification should be fully agnostic w.r.t. the internal representation used by the black-box; and $(ii)$ a good explanation language for graph classification tasks should be represented by motifs, as they are higher-order structures
which encode more information and are more adequate for human understanding. Following these observations, in this paper we introduce \method.

\spara{Contributions.} 
\method is a Shapley-based approach able to explain a black-box output assigning importance scores to a set of motifs. These input motifs can be provided by the domain expert or mined from a dataset (when available). This decoupling of the input and the explanation spaces is the main challenge and the main element of differentiation w.r.t. most explainability approaches.
In particular, we consider a specific task, named graph classification \emph{with node identity awareness} \cite{gutierrez2019embedding,you2021identity,lancianokdd,AbrateB21} that arises whenever a specific node id corresponds to the same entity in all the input networks: for instance, in classification of brain networks, the same node id represents the same brain region in all the input graphs. While not every application domain has node-identity awareness, it is crucial to exploit this property whenever it occurs, as ignoring it represents an important loss of information.
Our solution is model-agnostic: given a ``black-box'' $B$, we do not require to know the learning algorithm that produced $B$, nor its parameters, nor we have access to its training data. $B$ might as well be an opaque executable or an exposed API. The only requirement about $B$ is that it can be queried at will.
We also explore different attribution strategies and an approximation kernel to decrease the computational complexity of the approach. 

\spara{Roadmap.}
We first collocate our contribution in the literature~(\S \ref{sec:rw}); then we formalise the problem statement~(\S \ref{sec:problem_statement}) and introduce \method~(\S \ref{sec:pipeline}). We devise a synthetic dataset generator for graph classification which we use to benchmark \method and to validate the attribution strategies and the approximation kernel~(\S \ref{sec:experiments}).
Finally, we test \method on a real-world brain-network dataset consisting of patients affected by Autism Spectrum Disorder and a control group. Our experiments highlight how the classification provided by a black-box model can be effectively explained by few connectomics patterns (\S \ref{sec:brain}). \\
\section{Background and Related Work}
\label{sec:rw}
\spara{Explanations with Shapley values.}
SHAP~\cite{lundberg2017unified} derives local explanations leveraging the concept of \emph{Shapley values} from cooperative game theory \cite{shapley1953value,vstrumbelj2014explaining}.
In this context, the input features of a classifier are akin to players cooperating to win a game (the model prediction). The more important is a player to the cooperation, the higher is its Shapley value. Features are grouped into \textit{coalitional sets}, corresponding to the power set of the set of features $F$.
A \textit{coalitional game} is a function $v: 2^{|F|} \to \mathbb{R}$ that maps a coalitional set $S$ to a \textit{gain} value.
It follows that, for a feature $f_i \in F$, its Shapley value $\varphi(i)$ is defined as follows:
\begin{equation*}\label{eq:phi}
\varphi(i) = \sum_{S \subseteq F\setminus\{i\}}
\frac{\binom{\left|F\right|+1}{\left|S\right|}^{-1}}
{\left|F\right|+1}
(v(S \cup \{i\})\text{-}v(S)).
\end{equation*}
where $v(S)$ is the \textit{gain} value of a coalitional set and the sum extends over all subsets $S$ of $F$ not containing feature $i$.
SHAP attributes importance $\varphi_i$ to each feature $i \in F$. Importance is determined with respect to one (or more) \emph{background values}, which are typically chosen as the centroids of the dataset $X$. A feature is important if its deviation from the background value produces a large variation in the model output.
Since the exact evaluation of Shapley values requires to compute $2^{|F|}$
marginalisations that can result in a computationally infeasible process, several approaches were proposed to provide estimates of $\varphi_i$
by sampling only a fraction of the entire possible space of $S$ \cite{lundberg2017unified,sundararajan2020many,mitchell2021sampling}. However, when features $F$ are not independent, convergence towards the exact Shapley values can require to sample a higher number of configurations \cite{chen2020true}.

\spara{Explainable graph classification.}
A recent taxonomic survey \cite{yuan2022explainability} grouped the existing instance-level explanation techniques for graph neural networks in few classes: gradients/features-based~\cite{baldassarre2019explainability,pope2019explainability,schnake2020higher}, perturbation-based~\cite{luo2020parameterized,ying2019gnnexplainer,schlichtkrull2020interpreting,yuan2021explainability}, and surrogate methods~\cite{huang2020graphlime,zhang2021relex,vu2020pgm}.
Most of the above mentioned explanation methods have been devised for node classification and link prediction: these are different classification tasks from the one that we consider in this paper, i.e., classification of graphs with node identity, which has received much less attention.
While some of them seem a promising approach also for explaining graph classification models, their use has not yet been demonstrated for classification of graphs with node identity. Moreover, almost all explainers are unable to decouple their explanation language from the features space of the model, as such they are not suitable to provide explanations by means of motifs, and thus hard to compare with our proposal.


For instance, GNNExplainer~\cite{ying2019gnnexplainer} provides an importance score for each node and each edge: this fine-grained style of explanation might possibly overwhelm with scattered information the user seeking an explanation, moreover it is not clear how nodes scores and edges scores relate to each other, and it is thus hard to imagine how to use them to assign an importance to motifs.

Similarly to our proposal, SubgraphX~\cite{yuan2021explainability} considers a connected subgraph (or motif) as more comprehensible explanation than isolated edges/nodes. However, there are key differences between our approach and SubgraphX:
(i) in SubgraphX the explanation corresponds to a set of nodes and its induced subgraph, while \method computes attribution scores, and therefore relative numerical importance, for an arbitrary set of motifs, that can be mined or user-defined;
(ii) our approach is deterministic, while SubgraphX is based on MonteCarlo tree search; and (iii) \method can assign an importance score to any subgraph, spanning from a single edge to complex patterns, while SubgraphX is limited to the unique subgraph induced by a set of nodes.

\spara{Identity-aware graph ML.}
When it is possible to identify the same node among different graphs, it is crucial to take into account the identity of the node, as just ignoring it would represent a fatal loss of information. 
Beside brain network classification, which is our driving application,  this type of scenario is  very frequent in \emph{“omics”} data analysis where high-throughput measurements can be directly represented as networks (see ~\cite{biological_network} for a survey of networks representations of \emph{“omics”} data).
For instance, \emph{differential gene co-expression analysis} 
\cite{bth379} is an important approach aimed at identifying changes in response to an external perturbation, such as mutations predisposing to cancer development, and leading to changes in the activity of gene expression regulators or signaling. Given co-expression networks coming from different groups (e.g. patient and control), it is important to build classifiers and to explain them in terms of more complex structures such as motifs. The same problem arises in the context of e.g., \emph{protein-protein interaction networks} \cite{wsbm121,gulfidan2020pan}, or \emph{gene regulatory networks} \cite{kim2018diffgrn,singh2018differential}.
The increasing awareness about the importance of this topic is also mirrored by a growing number of ML models able to exploit node identity in graphs~\cite{gutierrez2019embedding,lancianokdd,you2021identity,AbrateB21}. In particular, \cite{you2021identity} includes a discussion about how the expressive power of identity-aware GNNs is superior to standard GNNs, showing how the expressive power of these models goes beyond the 1-Weisfeiler-Lehman (1-WL) graph isomorphism test.\\
\section{Problem Statement}
\label{sec:problem_statement}
We are given a graph $G = \mathcal{(V,E)}$ with nodes $i,j \in \mathcal{V}$, edges $(i,j) \in \mathcal{E}$.
We let $\mathcal{G}$ denote the set of all possible graphs defined over a fixed set of nodes $\mathcal{V}$. We are also given a black-box graph classification model $B$, that is, we can not inspect $B$, we do not know the learning algorithm that produced $B$, nor its parameters, nor we have access to
its training data. The only requirement about $B$ is that it can be queried at will. Without loss of generality we assume a binary classification with class labels $\{0,1\}$; in particular we model the black-box as a function $B: \mathcal{G} \to [0,1]$
returning, for any graph $G \in \mathcal{G}$, the probability that $G$ is in class 1.
For the scope of this paper, we consider $G$ to be an undirected, unweighted graph. In this context a motif $M$ is any \emph{connected} subgraph defined over any subset of $\mathcal{V}$. Formally, the set of all possible motifs is defined as
$$\mathcal{M} = \{
M_i = (\mathcal{V}_i,\mathcal{E}_i) | \mathcal{V}_i \subseteq \mathcal{V} \wedge \mathcal{E}_i \subseteq V_i \times V_i \wedge M_i \, \text{is connected}\}.$$

\begin{mdframed}[innerbottommargin=3pt,innertopmargin=3pt,innerleftmargin=6pt,innerrightmargin=6pt,backgroundcolor=gray!10,roundcorner=10pt]
Given a graph $G \in \mathcal{G}$, a black-box $B: \mathcal{G} \to [0,1]$ and a set of motifs $\mathcal{M}$, the problem tackled in this paper is that of assigning an \emph{explanation score} $\gsfunc(G,B,M_i) \in [-1,1]$ to each motif $M_i \in \mathcal{M}$, quantifying the impact of the motif in explaining the label $B(G)$: a value close to -1 means that $M_i$ is important in explaining $B(G) = 0$, a value close to 1 means that $M_i$ is important for $B(G) = 1$.
\end{mdframed}

\smallskip Moreover, when dealing with a large explanation language where the exact Shapley values calculation is computationally unfeasible, we tackle the problem of approximating the feature importance attribution by drastically shrinking the number of Shapley marginalisations, without compromising the correct estimation of the features importance.\\

\section{\method}
\label{sec:pipeline}

\begin{figure*}[t!]
  \centering
    \includegraphics[width=0.95\textwidth]{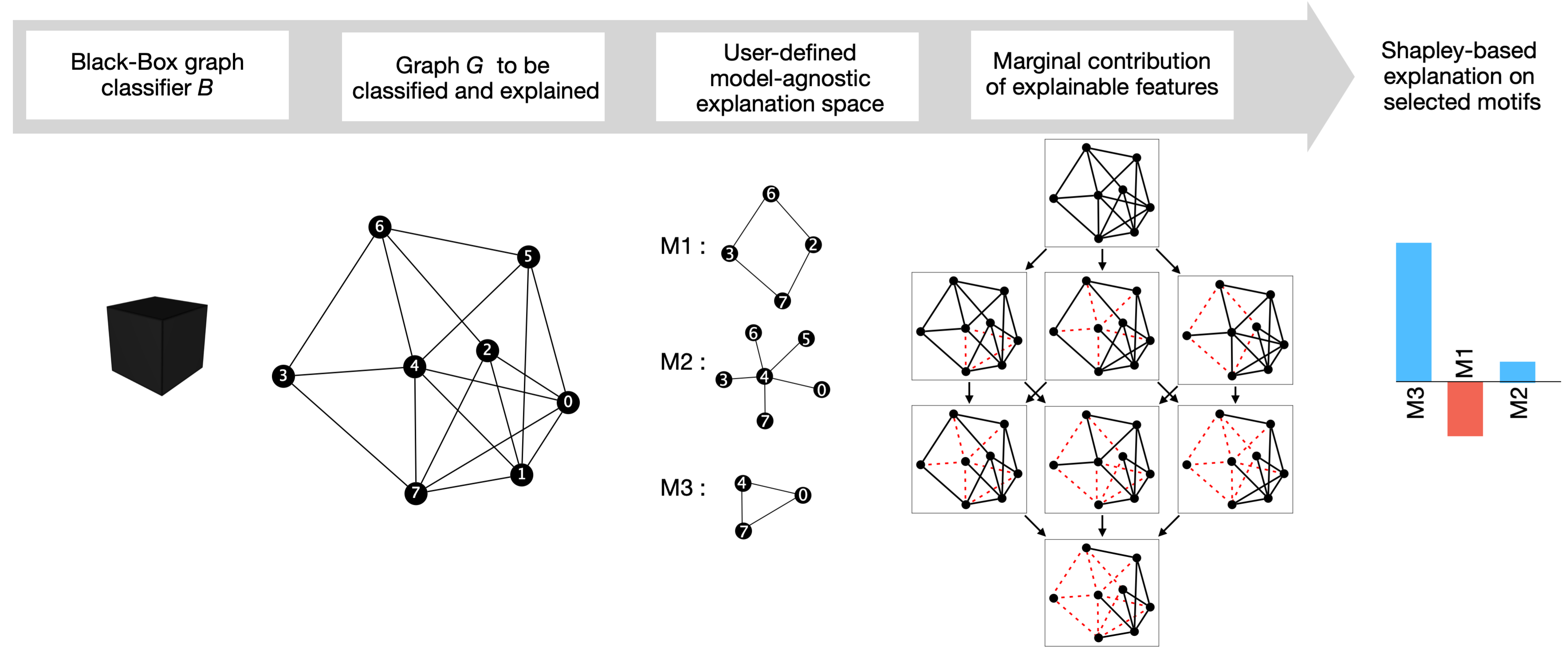}
    \caption{\method framework. The inputs are a black-box graph classifier, a graph whose classification by the black- box is to be explained, and an arbitrary set of motifs as an explanation language. \method leverages a lattice-based process of maskings and marginalisations to compute the explanation score of each motif, presenting a visual explanation to the user.}
    \label{fig:overview}
\end{figure*}

As previously stated, our approach is rooted in the computation of Shapley values as in~\cite{lundberg2017unified}, with the key difference that the explanation space does not need to correspond to the features $F$ (the input space of $B$); conversely, we base our explanations on the arbitrarily defined set of motifs $\mathcal{M}$. Mapping this approach and Equation~\ref{eq:phi} to our setting, we obtain the single-motif explanation score:
\begin{equation*}
\gsfunc(G,B,M_i) =
\sum_{S \subseteq \mathcal{M}\setminus\{M_i\}}
\frac{\binom{\left|\mathcal{M}\right|+1}{\left|S\right|}^{-1}}{\left|\mathcal{M}\right|+1}
{(B(G_S)\text{-}B(G_{S \cup \{M_i\}}))}.
\end{equation*}
The overall structure of the Shapley formula is preserved, with $v(\cdot)$ being replaced by $B(\cdot)$, as the coalitional value is provided by querying the black-box on a masked $G_S$, i.e., the graph $G$ from which the motifs in $S$ are ``masked".
$G_S$ is the graph-equivalent of the feature removal concept upon which the Shapley theory is based. In our setting we call this operation {\em motif masking} (discussed in detail in \S \ref{sec:motif_masking}). The subtraction operation is specular with respect to the original Shapley formulation because in our setting we are masking motifs rather than adding features: that is, if in standard Shapley the information about a feature $i$ is added by transitioning from $S$ to $S \cup \{i\}$, in our setting we obtain the same by transitioning from $G_{S \cup \{M_i\}}$ to $G_S$, by un-masking $M_i$.
Figure~\ref{fig:overview} summarizes our framework: given a graph $G$, a black-box classifier $B$, and a set of motifs $\mathcal{M}$, \method provides a motifs-based explanation $ \gsfunc(G, B, \mathcal{M})$. The set of motifs $\mathcal{M}$ might be directly provided by the domain expert as hypotheses to be tested or, in the case a dataset of graphs defined over the same $\mathcal{V}$ is available, interesting motifs (e.g., frequent or discriminative)  can be mined. We remark that the focus on motifs is an arbitrary choice, as we deem them to be a valuable and comprehensible explanation language; \method can also be applied as it is to investigate the importance of single edges and non-connected subgraphs.

\DontPrintSemicolon

\begin{algorithm}[h]
\small
 \KwIn{graph $G$, black-box $B$, set of motifs $\mathcal{M}$, \\masking strategy $ms$}
 \KwResult{explanation $\gsfunc(G,B,\mathcal{M})$}
\For{$S \subseteq \mathcal{M}$}{
Compute $B(G_S)$ according to $ms$;
}
$\gsfunc(G,B,\mathcal{M}) \leftarrow \emptyset$;\\
\For{$M_i \in \mathcal{M}$}{
$\gsfunc(G,B,M_i) \leftarrow 0$;\\
\For{$S \subseteq \mathcal{M}\setminus\{M_i\}$}{
$val \leftarrow B(G_S)-B(G_{S\cup\{M_i\}})$;\\
$coef \leftarrow \binom{\left|\mathcal{M}\right|+1}{\left|S\right|}$;\\
$\gsfunc(G,B,M_i) \leftarrow \gsfunc(G,B,M_i) + coef*val$;
}
$\gsfunc(G,B,\mathcal{M}) \leftarrow \gsfunc(G,B,\mathcal{M}) \cup \gsfunc(G,B,M_i)$;
}
\Return $\gsfunc(G,B,\mathcal{M})$
\caption{\method}
\label{algo:gshap}
\end{algorithm}

As for other Shapley-based approaches, \method (whose pseudocode is provided in Algorithm~\ref{algo:gshap}) requires to explore the powerset of the set features (which in our case is the set of motifs $\mathcal{M}$). This operation can be implemented by means of a lattice data structure. The first step (line 2) materializes all possible partial maskings $G_S$ of $G$ by exploring the lattice and then querying the black-box for every such partial masking. Identity awareness of the motifs enables to run this step efficiently, since each motif describes a set of edges between specific, unique nodes; other motifs with the same pattern of $M_i$ but involving different nodes are not involved in the masking of $M_i$.
The second step (line 8) involves computing the difference between scalar values of subsets of $\mathcal{M}$ which differ only by one motif; this difference quantifies the answer to the question {\em how would my classification score change if that specific motif was masked?}
Finally, in lines 9,10 we compute the marginal contributions of each motif by weighting the values obtained in line 8 with binomial coefficients, in order to account for the different number of edges across lattice layers. We then sum together all weighted values corresponding to the same motif, obtaining $\gsfunc(G,B,M_i)$ for all motifs $M_i$. We denote \method's $\mathcal{M}$-based explanation for the label assigned by $B$ to $G$ as $\gsfunc(G,B,\mathcal{M}) = \{\gsfunc(G,B,M_i),  \forall \ M_i \in \mathcal{M}\}$.

\subsection{Motif masking}
\label{sec:motif_masking}
Shapley values assess the relevance of a player by measuring the difference in gain between each coalition with/without that player: in the machine learning setting, this corresponds to
feature masking. As most of the classifiers cannot be trained on a dataset with a given number of features and then be queried over data-points with a different number of features, the common workaround is to replace each feature to be masked with its so-called {\em background value}, typically corresponding to the mean/median value for that feature in the dataset. 
In our context we do not need to substitute features with their background value:
we can always query the black-box classifier with any graph $G$, as far as it is defined over the same set of nodes  $\mathcal{V}$. As features are motifs, we can mask motifs from a graph, obtaining another graph that can be fed to the black-box. In the following we propose two simple motif masking approaches that we will later compare empirically in \S \ref{sec:masking_experiments}.

\spara{Remove.} The first motif masking approach simply removes from $G = (\mathcal{V},\mathcal{E})$ all edges occurring in any motif in $S$.
Formally, we define the resulting masked graph $G_{\mathcal{M}}$ as
$$G_{\mathcal{M}} = (\mathcal{V},\{e \in \mathcal{E} \mid e \not\in \mathcal{E}_i \ \forall \ M_i = (\mathcal{V},\mathcal{E}_i\mathcal{)} \in \mathcal{M}\}).$$


\spara{Toggle.} In an alternative strategy, each edge belonging to a motif to be masked is
removed from $G$ if occurring, and added otherwise. This masking strategy can be computed by symmetric difference:
$$G_{\mathcal{M}} = \mathcal{(V},\mathcal{E} \triangle\  (\bigcup_{\mathcal{M}_i = \mathcal{(V},\mathcal{E}_i) \ \in \ \mathcal{M}} \mathcal{E}_i)).$$
This masking strategy equally captures the information conveyed by the absence of a link and motif.

\subsection{Approximation kernel}
\label{sec:kernel}


The supporting data structure for the computation of Shapley values is a lattice, where each node represents which features are to be masked, and each edge represents a marginalisation with respect to a given feature. The Shapley value of a feature is the weighted sum of all marginalisations associated to that feature. 
Another important concept is the depth of a layer of edges - that is, the distance from the closest terminal nodes of the lattice (top or bottom). Given a set of features $F$, a node representing a coalition $S$ has depth $min(|S|, |F\setminus S|)$. Intuitively, depth-1 marginalisations represent the impact of single features, and in general the two depth-$k$ layer captures the impact of groups of $k$ features.
The approximation can be performed by random sampling the lattice nodes~\cite{vstrumbelj2014explaining},
but visiting nodes in a layered fashion
with increasing depth has been shown to create better approximations~\cite{lundberg2017unified}.
We follow this latest approach and compute approximations by summing marginalisations up to a given depth in the Shapley lattice.
We show experimentally that in our setting, i.e. when the explanation space differs from the input space, this approach provides an excellent estimate of the feature importance even when the motifs are not independent.
The depth-1 approximation allows to drastically reduce the operations from order $2^{|F|}$ to $2*\left|F\right|$.\\
\FloatBarrier 
\section{Experiments}
\label{sec:experiments}

\subsection{Synthetic dataset generation}
\label{sec:sythentic_data_generation}

Here we produce synthetic graph datasets with controlled motifs injection, in such a way that we can know, \emph{ex ante}, which motifs are more discriminative, and thus should have higher explanation score. In particular,
we produce a set of synthetic graphs $\mathcal{G}$ with labeling  $\tau: \mathcal{G} \to \{0,1\}$. A parameter $\rho(k)$ controls how much each injected motif $M_k$ is predictive of the mapping $\tau$, as follows. We consider a fixed set of nodes $\mathcal{V} = \{1,...,n\}, n \in \mathbb{N}$, and a set $\mathcal{M}$ of $m$ motifs defined over  $\mathcal{V}$, such that each motif $M_k \in \mathcal{M}$ is predictive of class $0$ if $k$ is even and of class $1$ otherwise.
A correlation matrix $\mathbf{C} \in [0,1]^{m \times m}$ is given as input to control if motifs are added independently or not: $c_{i,j}$ represents the probability that motif $j$ is added to a graph, given that motif $i$ was added.
We create $\mathcal{G} = \{G_0, ..., G_{|\mathcal{G}|-1}\}$ with Erdős–Rényi (ER) graphs having density $\nu$ and mapping function $\tau(G_i) = 0$ if $i$ is even and $\tau(G_i) = 1$ otherwise.
Then, for each graph-motif pair ($G_i$,$M_k$), with a \emph{perturbation probability} proportional to $\rho(k)$
we add the motif to the graph if motif and graph belong to the same class, and remove it otherwise.


For this experiment we build a synthetic dataset $(\mathcal{G},\tau)$ with 250 graphs per class defined over $|\mathcal{V}|=100$ nodes, with density $\nu = .2$ and with 6 injected motifs all containing 10 edges. The 6 motifs have increasing perturbation probability values $0,0.15,0.3,0.45,0.6,0.75$, spanning from $M_0$ being a random control motif, to $M_5$ having being $75\%$ probability of being injected (resp. removed) in graphs of class 1 (resp. class 0). As explained above, $M_i$ is predictive of class 0 if $i$ is odd, and of class 1 otherwise. The correlation matrix between the motifs is set to the identity, meaning that the motifs are chosen independently. Once the dataset $(\mathcal{G},\tau)$ is defined, the next step is to train a ML model to obtain a black-box graph classifier $B$.
We found that the following model architecture has good performance in predicting the graph class: a GNN where the first layer is a graph transformer~\cite{unimp_ijcai21} followed by batch normalization~\cite{ioffe2015batch}, then another graph transformer followed by batch normalization, followed by two fully connected layers.
In this architecture, node identity can be enforced through the use of the identity as feature matrix. The final accuracy score on the test set is 0.79.

Finally, we compute the \method scores for all graphs. Spanning from $M_0$ to $M_5$, the array of average motif scores are $[0.0009,  0.001,  -0.0049,  0.0072, -0.0108,  0.0263]$. The alternation between positive and negative values mirrors the parity-based criterion we used to assign motifs to classes: odd-indexed motifs have positive explanation scores, and the even-indexed ones have negative explanation scores. Furthermore, since these motifs were injected with increasing probabilities, this is mirrored by the monotonically increasing absolute values of our explanations scores.

\subsection{Comparing motif-masking strategies}
\label{sec:masking_experiments}

\begin{figure}[t!]
\centering
\begin{subfigure}{.49\textwidth}
  \centering
  \includegraphics[width=\linewidth]{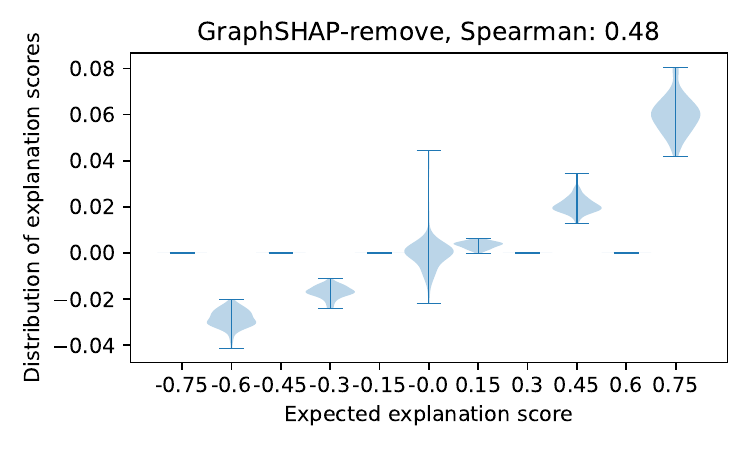}
\end{subfigure}
~
\begin{subfigure}{.49\textwidth}
  \centering
  \includegraphics[width=\linewidth]{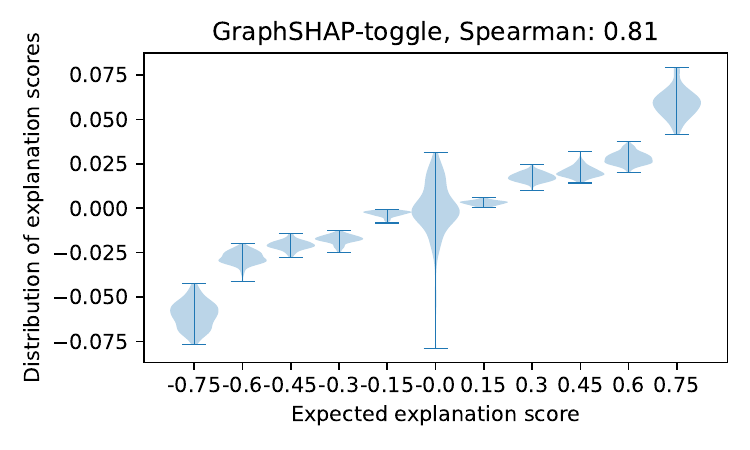}
\end{subfigure}
\caption{Distribution of explanation scores produced by \method, given an expected explanation score using the {\em remove} (left) and {\em toggle} (right) masking strategies.}
\label{fig:violin_gshap}
\end{figure}
Using the same experimental setting of the previous section, we compare the different motif-masking strategies
{\em remove} and {\em toggle}. In Figure~\ref{fig:violin_gshap} we show the distribution of $\gsfunc$ scores ($y$-axis) for different values of \emph{expected explanation score} ($x$-axis, defined next).
Expected explanation scores are obtained by considering the following:
(1) motifs that are injected in graphs of class $1$ should have positive score, while motifs that are removed should have negative score;
(2) a motif that is neither injected or removed from $G$ should have low absolute score value;
(3) the absolute score value is proportional to how much predictive is the motif for the black-box, thus it should be proportional to the probability of injection $\rho$.
Therefore the expected explanation score is defined as $I(G_i,M_j) \cdot C_{M_j} \cdot \rho(M_j)$, where $I(G_i,M_j)$ is $1$ if $M_j$ was injected in $G_i$,  $-1$ if $M_j$ was removed from $G_i$, $0$ otherwise, and $C_{M_j}$ is $1$ if the class associated to motif $M_i$ is $1$, $-1$ otherwise.
Figure~\ref{fig:violin_gshap} shows that both strategies produce explanation scores that display an increasing trend matching the expected score, and in particular \textit{toggle} best correlates (Spearman coefficient of 0.81) with the expected explanation given these assumptions.

\subsection{Comparison with similar explainers}

\begin{figure}[t!]
\centering
\begin{subfigure}{.49\textwidth}
  \centering
  \includegraphics[width=\linewidth]{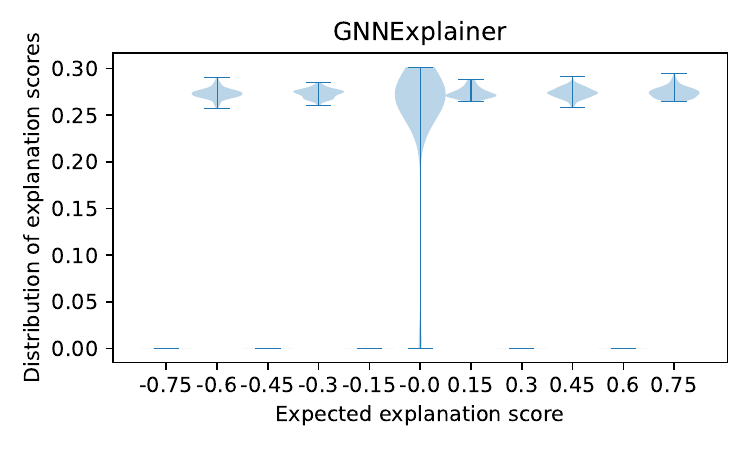}
\end{subfigure}%
~
\begin{subfigure}{.49\textwidth}
  \centering
  \includegraphics[width=\linewidth]{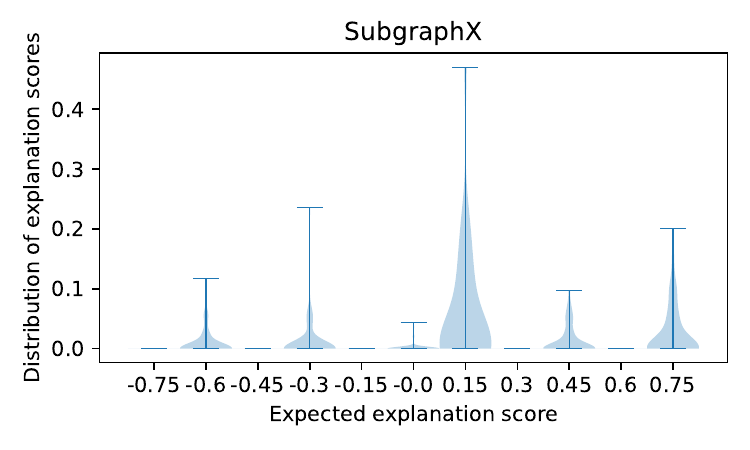}
\end{subfigure}
\caption{Distribution of explanation scores produced by GNNExplainer (left) and SubgraphX (right), given an expected explanation score.}
\label{fig:violin_others}
\end{figure}

As stated in \S \ref{sec:rw}, \method has the unique feature of providing explanations for a user-defined set of motifs. However, for the sake of comparison, we include a brief discussion of the differences with two relatively similar explainers, GNNExplainer~\cite{ying2019gnnexplainer} and SubgraphX~\cite{yuan2021explainability}; we use the same experimental settings of the previous section, results are shown in Figure~\ref{fig:violin_others}.

GNNExplainer returns a {\em mask} with edge-level scores. In order to correlate this explanation language with the expected motif explanation scores, we assign to each injected motif the average of the GNNExplainer scores for the arcs of that motif. We remark that, by construction, GNNExplainer can only identify edges that have a positive impact on the black-box classification outcome.
SubgraphX return a single set of nodes, so that the produced explanation is the subgraph induced by that set of nodes. In order to correlate this explanation with our motif-based expected explanation scored, for each graph we computed the six Jaccard similarities between the single SubgraphX-produced motif and the six injected motifs.
We observe how both explainers are unable to capture the different importance between frequent and rarely-injected motifs, or to assign importance to the absence of edges. We remark that the introduced synthetic graph benchmark is motif-based in order to validate \method; nonetheless, it allows for discussion about the differences with other explainers, especially in terms of explanation languages.

\subsection{Depth-based approximation}
\label{sec:kernel_experiments}

To test our depth-based approximation strategy for the computation of Shapley values, we generated different datasets, computed approximations at all depths, and compared the approximated results with the exact explanation scores for all cases. We observed that the Shapley scores are strongly correlated with the marginalisations that are calculated by just taking into account a single motif $M_i$ or the coalition of $\mathcal{M} - M_i$, i.e. the most external marginalisations of the lattice. This correlation could be due to the fact that the insertion of motifs are completely independent of each other. To test the hypothesis that Shapley scores can be in large part approximated by the external marginalisations, independently of motif correlations, we created an initial dataset with 5,000+500 (train/test) graphs, 100 nodes per graph, initial graph density of $0.2$, 8 motifs, perturbation probability of $0.25$, and no forced correlations between motifs. 
We then forced partial and total correlation between a couple of same-class motifs ($\mathcal{M}_0$ and $\mathcal{M}_2$) by inserting the values of $0.5$ and $1$ in the correlation matrix - so that, for instance, partial correlations was obtained by setting $\mathbf{C}_{0,2} = \mathbf{C}_{2,0} = 0.5$, where 0 and 2 are the indexes of the two selected motifs. We also repeated the same process with a triplet of same-class motifs $\mathbf{C}_{0,2} = \mathbf{C}_{2,0} = \mathbf{C}_{0,4} = \mathbf{C}_{4,0} = \mathbf{C}_{2,4} = \mathbf{C}_{4,2} = 0.5$ (or $= 1$ for strong correlation).

\begin{figure}
\centering
\begin{subfigure}{.49\textwidth}
  \centering
  \includegraphics[width=\linewidth]{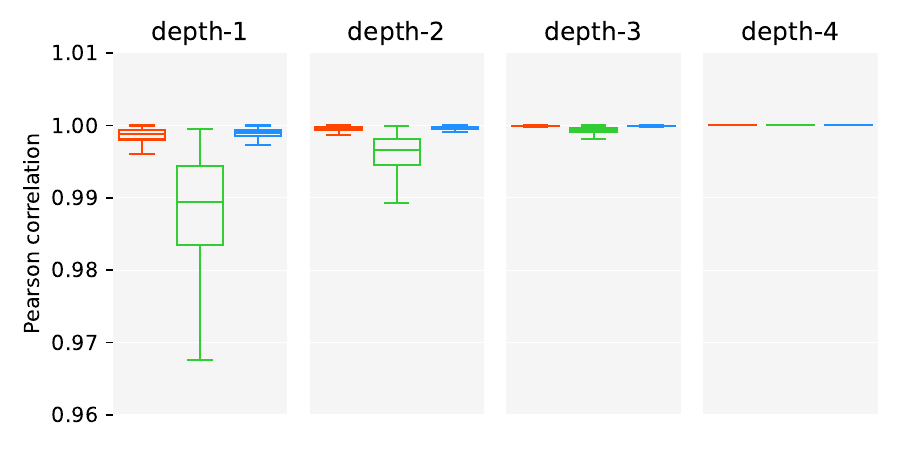}
  \caption{Pairwise correlation}
\end{subfigure}%
~
\begin{subfigure}{.49\textwidth}
  \centering
  \includegraphics[width=\linewidth]{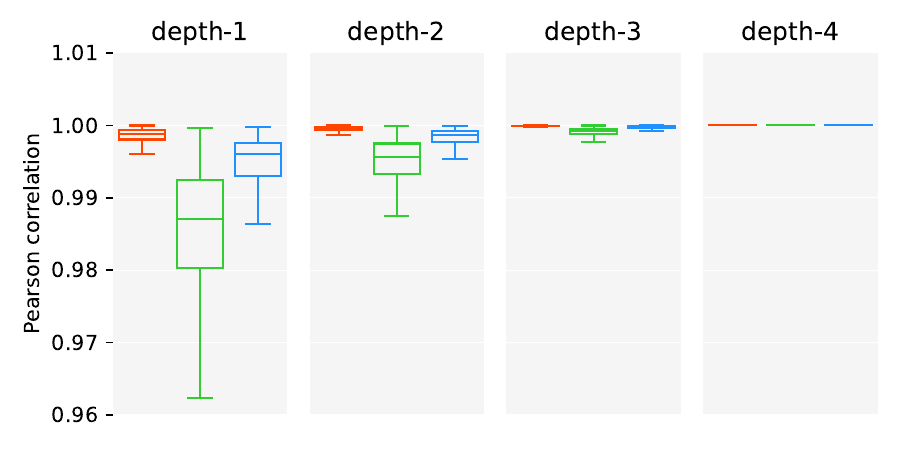}
  \caption{Triplet correlation}
\end{subfigure}
\caption{Pearson correlation between exact and approximated explanation scores. Explanation scores were approximated in datasets with no correlation (red), partial correlation (green) or strong correlation (blue).}
\label{fig:kernel}
\end{figure}

As mentioned above, for all five datasets we run the whole \method pipeline, training a GNN with the same architecture as \S\ref{sec:sythentic_data_generation} as black-box, obtaining a test accuracy of $0.8 \pm 0.02$ in all cases.
We computed the Shapley-based explanation scores for each graph in the test set. We then computed all depth-based approximations: with 8 motifs, the approximation depth spans from 1 to 4, included. We also remark that a depth-4 approximation takes into account all depths and therefore corresponds to the exact explanation scores. We compared each approximation with the exact explanation scores by means of Pearson correlation to measure how much approximations can portray the relative importance of the exact explanation scores.

The results of this experiment are reported in Figure~\ref{fig:kernel}. Each boxplot represents the distribution of Pearson correlation scores over the test set of 500 graphs, for a given approximation depth and feature correlation level. As expected, the higher the approximation depth, the better the explanation values are approximated to the exact values. While the results show some fluctuations amongst the datasets,
we observe that in all depth-1 scenarios the correlation scores are above 0.95, meaning that depth-1 approximated explanation scores already achieve a very good approximation to the exact feature importance. This is remarkable, given that depth-1 approximation requires two marginalisations per feature, while the computation of exact explanation scores requires to perform an exponential number of marginalisations.\\


%
\section{\method on brain data}
\label{sec:brain}
In this section we apply \method on a brain network dataset to show how a black-box for the diagnosis of Autism Spectrum Disorder can be explained with a custom set of motifs.
We used the publicly-available {\em eyesclosed} dataset released by the Autism Brain Imagine Data Exchange (ABIDE) project~\cite{abide}. The dataset contains neuroimaging data of 871 different patients, 468 Typically Developed (TD, class $0$) and 403 suffering from Autism Spectrum Disorder (ASD, class $1$). We performed standard data pre-processing\footnote{Code from \href{https://github.com/parisots/population-gcn}{population-gcn} and \href{https://github.com/largeapp/gat-li}{gat-li}}, obtaining graphs defined over 110 nodes corresponding to the ROIs of the Harvard Oxford atlas \cite{desikan2006automated}; for each graph we kept the $10\%$ most significative pairwise interaction between ROIs. As black-box $B$ we adopt the same architecture as described in \S\ref{sec:experiments}; we re-implemented it and obtain the same test accuracy of 0.685. 
We extracted a set $\mathcal{M}$ of 10 interesting motifs from the dataset, to be used as explanation language (as discussed in \S\ref{sec:pipeline}, motifs could be provided by a domain expert or be generated in any other way).


\begin{figure}[h!]
\centering
\begin{subfigure}{.75\textwidth}
  \centering
  \includegraphics[width=\linewidth]{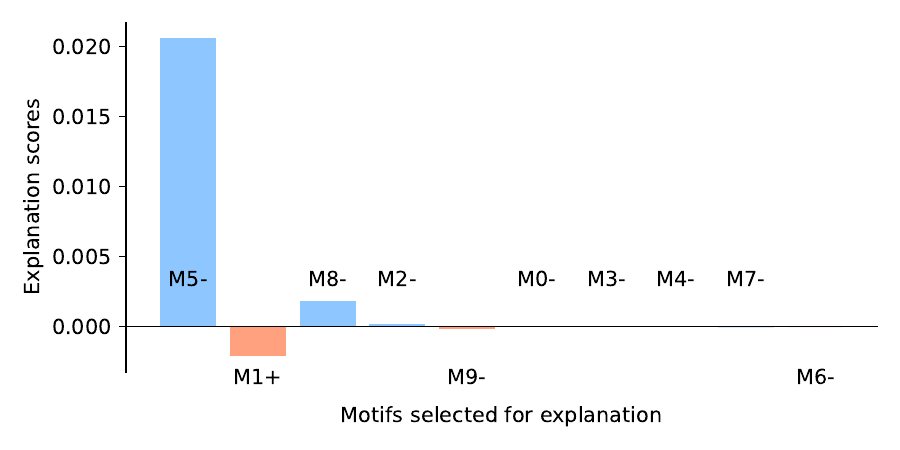}
\end{subfigure}%

\begin{subfigure}{.75\textwidth}
  \centering
  \includegraphics[width=\linewidth]{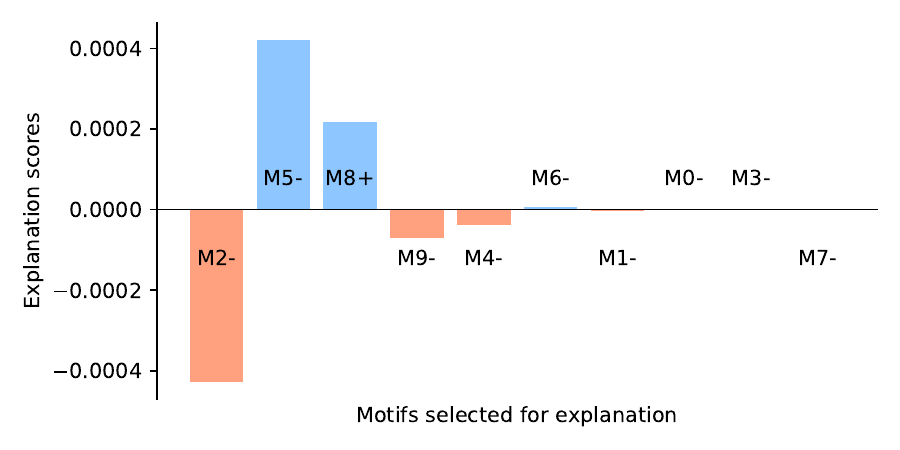}
\end{subfigure}
\caption{\method: local explanation for two single patients, $P_1$ (above) and $P_2$ (below)}
\label{fig:brain_local}
\end{figure}

We next apply \method (with {\em toggle} masking) to produce explanations based on $\mathcal{M}$ for the classifications produced by $B$.
Figure~\ref{fig:brain_local} shows an example of  $\mathcal{M}$-based \method explanation for two ASD patients, which will be denoted as $P_1$ and $P_2$ for brevity. At a glance, we can observe that $P_2$ has a more complex explanation, while $P_1$'s decision was pretty much due to a single motif. Furthermore, observing the $y$ axis, we note that the maximum score for $P_1$ is around $.02$, while for $P_2$ is barely $.0004$: this tells us that $B$ was more confident when diagnosing $P_1$, and not as much for $P_2$. Note that, in Figure~\ref{fig:brain_local}, we have enriched each motif with a $+$/$-$ symbol to denote whether the motif occurs in the graph associated to the patient. $P_1$'s diagnosis was justified by the {\em absence} of $M_5$: $M_5$ does not occur in $P_1$'s associated graph, and \method assigned this a strong positive explanation score. This is also true for $P_2$, but for that diagnosis there is also a strong driver against the classification of ASD: the absence of $M_2$. Finally, it is worth noting how both explanations assign positive explanation scores to $M_8$, despite it occurring in $P_2$ and not in $P_1$.

\begin{figure}[t!]
\centering
\includegraphics[width=.49\textwidth]{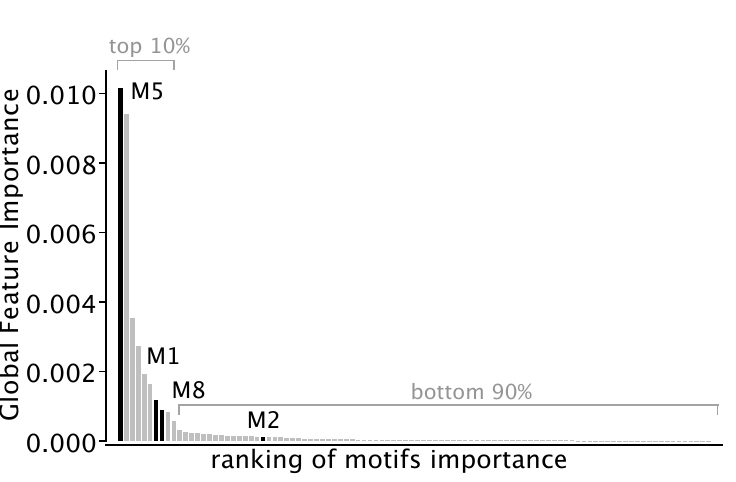}
~
\includegraphics[width=.49\textwidth]{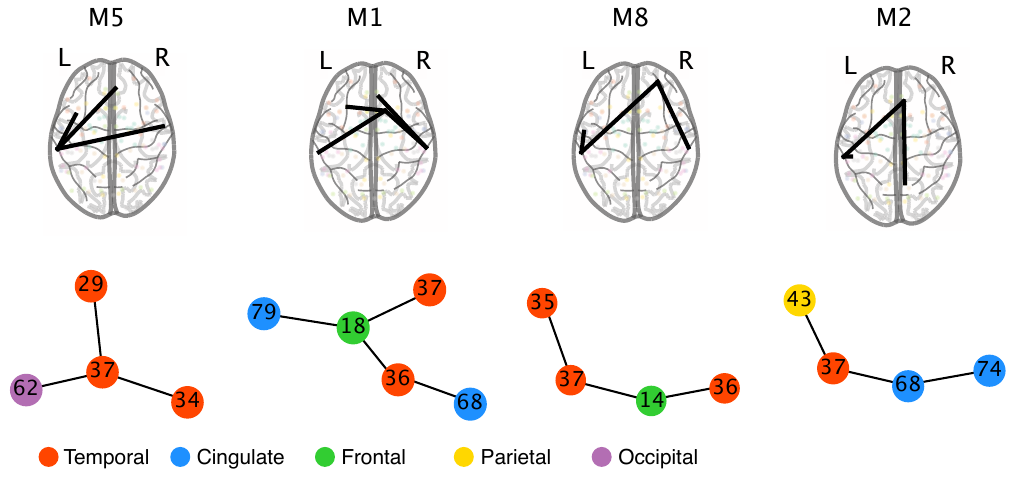}
\caption{Global explanation scores of 100 mined motifs, and brain mapping of the most relevant motifs of Figure\ref{fig:brain_local}}
\label{fig:brain_global}
\end{figure}

\method produces {\em local}, data-point-level explanations; it is therefore possible to collect explanation scores across the whole dataset and aggregate them to obtain a {\em global} feature ranking.
In \S \ref{sec:kernel_experiments} we observed how, for our synthetic datasets, a depth-1 approximation yielded high Pearson correlation scores with respect to the exact values. We run the same experiments for the brain dataset (for the 10 mined motifs of the previous section), obtaining analogous results: depth-1 approximations are extremely good proxies for the exact explanation scores (median Pearson's correlation coefficient across the test set is 0.997 and lower/upper quartile values are 0.994 and 0.999 respectively). 
Exploiting this result, we then extend the set of motifs, from the top-10 to the top-100 emerging from the ranking process, and compute their depth-1 approximated explanation scores. We remark that this process has linear computational cost with respect to the number of features, and therefore requires only 10x time with respect to the previous 10-features explanations. It is worth stressing that our approach allows to save $~10^{26}$ operations, making features attribution feasible in such a large space.
In Figure~\ref{fig:brain_global} we report, for each motif in $\mathcal{M}$, the average absolute value of the explanation scores across all graphs in $\mathcal{G}$. We observe that the average score drops rapidly, indicating that only few motifs are generally relevant across explanations. The motif with the highest global score is $M_5$, which we observed in both $P_1$ and $P_2$ in Figure~\ref{fig:brain_local}. It is also worth observing that $M_2$, a highly relevant motif for the explanation of $P_2$, has negligible global importance.
When a classifier is not inspected by a domain expert to validate existing hypothesis (i.e. asking explanations considering a small set of plausible motifs), the proposed kernel can be used to scale massively the number of explanations to generate novel hypothesis.
Conversely, further analysis of peculiarities such as $M_2$ might help the practitioner understand the decision process of the ML model.\\

\section{Conclusions and Future Work}
Although machine learning models aimed at classifying entire networks have important applications, 
suitable approaches for explaining such models are still missing. Here, we have presented \method, a novel approach for explaining black-box models for identity-aware graphs classifiers by means of expressive and understandable network motifs. 


Although \method builds explanations based on motifs, the approach can be expanded so that the explanation language includes additional network features, such as, e.g., nodes centrality, dense subgraphs, higher-order structures, or just groups of disconnected edges. This is a promising avenue for future investigation.
Finally, ongoing collaboration with neuroscientists will validate the usability of \method in validating neuroscience hypotheses connecting brain network motifs and neurodegenerative diseases.

All code is available at \href{https://github.com/alanturin-g/GraphShap_IJCNN2023}{this GitHub repository.}\\

\bibliographystyle{IEEEtran}
\bibliography{biblio}

%

\end{document}